\documentclass[conference,letter]{IEEEtran}  % Comment this line out if you need a4paper

\usepackage{amsmath,amsfonts}
\usepackage{algorithmic}
\usepackage{algorithm}
\usepackage{array}
\usepackage[caption=false,font=normalsize,labelfont=sf,textfont=sf]{subfig}
\usepackage{textcomp}
\usepackage{stfloats}
\usepackage{url}
\usepackage{verbatim}
\usepackage{graphicx}
\usepackage{cite}
\usepackage{subcaption}

\usepackage{amssymb}
\usepackage{orcidlink}
\usepackage{overpic}
\usepackage{graphicx} % for pdf, bitmapped graphics files
\usepackage{tabularx}
\usepackage{longtable}
\usepackage{booktabs}
\usepackage{lscape}
\usepackage{multirow}
\usepackage[table]{xcolor}

\definecolor{best}{gray}{0.75}
\definecolor{second}{gray}{0.92}
\newtheorem{theorem}{Theorem}[section]

\usepackage{tikz}

\usepackage{xcolor}

\usetikzlibrary{arrows.meta, fit, positioning, backgrounds, calc}

\definecolor{Cactor}{RGB}{30, 100, 185}
\definecolor{Ccritic}{RGB}{195, 95, 15}
\definecolor{Cenv}{RGB}{28, 135, 65}
\definecolor{Cbuf}{RGB}{105, 55, 175}
\definecolor{Calgo}{RGB}{55, 55, 65}
\definecolor{Cpd}{RGB}{175, 40, 85}
\definecolor{Cpanel}{RGB}{228, 238, 252}
\definecolor{Cpborder}{RGB}{155, 185, 225}

\tikzset{
  blk/.style={rectangle, rounded corners=4pt, minimum width=#1,
    minimum height=0.72cm, draw, very thick, align=center},
  blk/.default=2.4cm,
  actor/.style  ={blk=2.4cm, fill=Cactor!13, draw=Cactor,
                  font=\small\bfseries, text=Cactor},
  critic/.style ={blk=2.7cm, fill=Ccritic!13, draw=Ccritic,
                  font=\small\bfseries, text=Ccritic!90!black},
  envblk/.style ={blk=2.5cm, fill=Cenv!13,    draw=Cenv,
                  font=\small\bfseries, text=Cenv!90!black},
  bufblk/.style ={blk=2.7cm, fill=Cbuf!13,    draw=Cbuf,
                  font=\small\bfseries, text=Cbuf!90!black},
  algoblk/.style={blk=2.4cm, fill=Calgo!10,   draw=Calgo,
                  font=\small\bfseries, text=Calgo},
  iobox/.style  ={blk=2.1cm, fill=white, draw=gray!55, rounded corners=3pt,
                  font=\scriptsize\bfseries, text=gray!65!black,
                  minimum height=0.58cm},
  pdblk/.style  ={blk=1.7cm, fill=Cpd!10, draw=Cpd!75,
                  font=\scriptsize\bfseries, text=Cpd!80!black,
                  minimum height=0.58cm},
  intblk/.style ={blk=1.45cm, fill=Cpd!7, draw=Cpd!55,
                  font=\scriptsize\bfseries, text=Cpd!75!black,
                  minimum height=0.58cm},
  lbl/.style    ={font=\scriptsize, text=gray!65!black},
  A/.style ={-{Stealth[length=5pt,width=3.2pt]}, thick},
  Ab/.style={A, Cactor!80},
  Ag/.style={A, Cenv!80},
  Ao/.style={A, Ccritic!90!black},
  Ap/.style={A, Cbuf!80},
  Ad/.style={A, dashed, gray!55},
}

\begin{document}

\title{Hybrid TD3: Overestimation Bias Analysis and Stable Policy Optimization for Hybrid Action Space}

\author{
\IEEEauthorblockN{Thanh-Tuan Tran\textsuperscript{1}$~\orcidlink{0009-0006-4370-8326}$, Thanh Nguyen Canh\textsuperscript{2}~\orcidlink{/0000-0001-6332-1002}, Nak Young Chong\textsuperscript{2,3}~$\orcidlink{0000-0001-5736-0769}$ and  $^*$Xiem HoangVan\textsuperscript{1}$~~\orcidlink{0000-0002-7524-6529}$}
\IEEEauthorblockA{\textsuperscript{1}University of Engineering and Technology, Vietnam National University, 10000, Hanoi, Vietnam. \\ (\tt \{22023506, xiemhoang\}@vnu.edu.vn)}
\IEEEauthorblockA{\textsuperscript{2}School of Information Science, Japan Advanced Institute of Science and Technology, Nomi, 923-1211, Ishikawa, Japan. \\ (\tt\{thanhnc, nakyoung\}@jaist.ac.jp)}
\IEEEauthorblockA{\textsuperscript{3}Department of Robotics, Hanyang University, Ansan, 15588, Gyeonggi, Korea.}
\IEEEauthorblockA{*Corresponding author: \tt xiemhoang@vnu.edu.vn}
}

% The paper headers
\markboth{Journal of \LaTeX\ Class Files,~Vol.~14, No.~8, August~2021}%
{Shell \MakeLowercase{\textit{et al.}}: A Sample Article Using IEEEtran.cls for IEEE Journals}

\maketitle

\makeatletter
\setlength{\@fptop}{0pt}
\makeatother

%%%%%%%%%%%%%%%%%%%%%%%%%%%%%%%%%%%%%%%%%%%%%%%%%%%%%%%%%%%%%%%%%%%%%%%%%%%%%%%%
\begin{abstract}
Reinforcement learning in discrete-continuous hybrid action spaces presents fundamental challenges for robotic manipulation, where high-level task decisions and low-level joint-space execution must be jointly optimized. Existing approaches either discretize continuous components or relax discrete choices into continuous approximations, which suffer from scalability limitations and training instability in high-dimensional action spaces and under domain randomization. In this paper, we propose Hybrid TD3, an extension of Twin Delayed Deep Deterministic Policy Gradient (TD3) that natively handles parameterized hybrid action spaces in a principled manner. We conduct a rigorous theoretical analysis of overestimation bias in hybrid action settings, deriving formal bounds under twin-critic architectures and establishing a complete bias ordering across five algorithmic variants under synchronized Gaussian error assumptions. Building on this analysis, we introduce a weighted clipped Q-learning target that marginalizes over the discrete action distribution, achieving equivalent bias reduction to standard clipped minimization while improving policy smoothness. Experimental results demonstrate that Hybrid TD3 achieves superior training stability and competitive performance against state-of-the-art hybrid action baselines. Likewise, we show the consistent performance across five morphologically diverse robot platforms without platform-specific tuning further confirms that the stability properties of Hybrid TD3 are not contingent on specific kinematic structure.
\end{abstract}

% \begin{IEEEkeywords}
% Discrete–continuous actions, Reinforcement learning, Overestimation bias, Robotic manipulation.
% \end{IEEEkeywords}

%%%%%%%%%%%%%%%%%%%%%%%%%%%%%%%%%%%%%%%%%%%%%%%%%%%%%%%%%%%%%%%%%%%%%%%%%%%%%%%%
\section{Introduction}
\label{sec:introduction}

Hybrid action spaces are particularly attractive for robotic manipulation. Discrete modes naturally represent distinct operational behaviors, while continuous parameters specify the precise execution of each mode in the joint space. In practice, an agent must first decide \emph{what to do} before determining \emph{how to do it precisely}. Learned policies must therefore bridge high-level task decisions and low-level actuator commands simultaneously, operating directly in joint space to handle contact dynamics, joint limits, and scene uncertainty end-to-end~\cite{kumar2021joint}. On the other hand, policies might overfit to simulator-specific dynamics and are unlikely to transfer to the physical variation inevitable in real deployment without randomization. That makes domain randomization a necessary stress test for any policy intended for hardware use.

Conventional RL algorithms are designed for purely discrete or purely continuous action spaces. Common extensions to hybrid settings involve discretizing continuous components~\cite{hausknecht2016half}, parameterizing action subspaces~\cite{hausknecht2015deep}, or relaxing discrete choices into continuous approximations~\cite{massaroli2020neural}. These conversions suffer from scalability limitations: discretization leads to exponential action blowup, while relaxation introduces approximation errors that degrade policy quality. More principled approaches~\cite{hausknecht2015deep, xiong2018parametrized} directly model hybrid actions but overlook dependencies between discrete and continuous components, limiting convergence in high-dimensional settings. SAC-based hybrid variants~\cite{lin2024discretionary, delalleau2019discrete, xu2023action} improve exploration but introduce redundancy in parameter modeling that exacerbates instability under domain randomization.

The central difficulty is that the hybrid action structure and full domain randomization interact adversarially in the learning dynamics. A critical and underexplored source of this instability is overestimation bias. In standard Q-learning, the maximization operator over noisy value estimates systematically inflates target values~\cite{hasselt2010double, turcato2025exploiting}. Since this bias is compounded, the agent must simultaneously maximize over a discrete set of modes and optimize continuous parameters in a hybrid setting. Under full domain randomization, where object poses, masses, and scene configurations are resampled at every episode reset, the critic must regress a value function over a mixture of environment configurations drawn from a contextual distribution. This prevents reliable convergence of the critic and amplifies the bias feedback. A policy that fails to control this instability not only converges more slowly but also learns systematically incorrect discrete mode preferences early in training, biasing the entire subsequent trajectory of policy improvement. Existing hybrid action methods \cite{massaroli2020neural, xiong2018parametrized, lin2024discretionary, delalleau2019discrete, xu2023action, hasselt2010double} address structural challenges of hybrid spaces but do not analyze or bound overestimation bias under domain randomization.

To address these challenges, we first conduct a systematic empirical comparison of TD3~\cite{fujimoto2018addressing}, SAC~\cite{haarnoja2018soft}, DDPG~\cite{lillicrap2015continuous}, and PPO~\cite{schulman2017proximal}, each adapted to hybrid action spaces under full domain randomization. TD3 consistently achieves superior stability and performance, owing to its twin-critic architecture, delayed policy updates, and deterministic policy structure, which jointly mitigate overestimation bias and gradient variance. Motivated by this finding, we extend TD3 to hybrid action spaces and introduce a weighted clipped Q-learning target that marginalizes the Bellman backup over the full discrete action distribution, rather than committing to a greedy discrete selection. We further provide a rigorous theoretical analysis establishing a complete ordering of expected estimation bias across five hybrid algorithmic variants, formally justifying clipped double Q-learning as the most stable bias mitigation strategy under dense rewards and heavy randomization.
Our main contributions can be summarized as follows:
\begin{itemize}
    \item A systematic empirical analysis identifying TD3 as 
    the most stable backbone for hybrid action RL under full 
    domain randomization, with a formal characterization of 
    why competing baselines fail. We further validate this 
    stability across five morphologically diverse robot 
    platforms (UF850, KUKA, Panda, Kinova, UR5e), 
    demonstrating consistent high success rates without 
    platform-specific hyperparameter tuning, a practical 
    robustness property absent from prior hybrid action RL 
    evaluations.
    \item A weighted clipped Q-learning target for hybrid 
    action spaces that marginalizes over the discrete action 
    distribution, improving gradient smoothness while 
    preserving the bias properties of standard TD3.
    \item A rigorous theoretical bias ordering across five 
    hybrid algorithmic variants, providing principled guidance 
    for algorithm selection in hybrid action RL.
\end{itemize}
%===============================================================================
%===============================================================================

\begin{figure*}[!ht]
    \centering
    \begin{overpic}[width=\textwidth, unit=1pt]{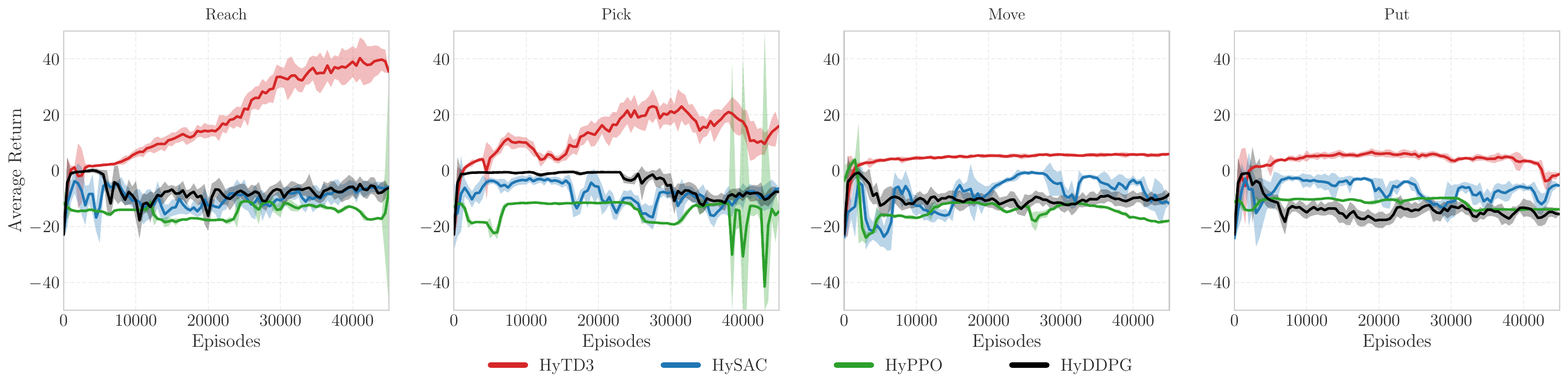}
        % \put(7.4, 3.8){\footnotesize Episodes}
    \end{overpic}
    \caption{HyTD3 exhibits superior reward performance than HySAC, HyPPO, and HyDDPG under domain randomization.}
    \label{fig:baselines}
\end{figure*}

\section{Methodology}
\label{sec:proposed}
This section presents our approach to learning hybrid action policies for robotic manipulation under full domain randomization. We begin by formulating the discrete-continuous hybrid action space and motivating the choice of TD3 as the most stable backbone through empirical baseline comparison. We then introduce our proposed modification — a weighted clipped Q-learning target and provide a theoretical analysis of overestimation bias across five hybrid algorithmic variants. Finally, we describe the state representation, and training protocol adopted for our manipulation tasks.

\subsection{Problem formulation}
We consider the robot manipulator setting in which an agent must make decisions spanning both discrete actions (\textit{e.g.} grasp, release) and continuous joint-space parameterization (\textit{e.g.}, joint velocities). This structure defines a \emph{parameterized hybrid action space}, where an action
$a$ combines a discrete component $a_d \in \mathcal{A}_d = \{0,1\}$
(selecting the suction mode) and a continuous component
$a_c \in \mathcal{A}_c \subseteq \mathbb{R}^{m_k}$ (mode-specific joint-space
parameters, where the dimensionality $m_k$ may vary per mode $k$). We adopt a factorized policy that models the two action components independently:
\begin{equation}
% \pi(a \mid s) = \pi_d(a_d \mid s) \pi_c(a_c \mid s)=\prod_{a^i \in a_d} \pi_d(a^i \mid s) \prod_{a^i \in a_c} \pi_c(a^i \mid s),
\pi(a \mid s) =\prod_{a^i \in a_d} \pi_d(a^i \mid s) \prod_{a^i \in a_c} \pi_c(a^i \mid s),
\end{equation}
where the product factorizes over independent discrete and continuous action components. The discrete part $\pi^d$ is parameterized as a categorical distribution implemented via a softmax over logits from a shared network trunk, and the continuous part $\pi^c$ is a deterministic actor head producing joint velocities. This factorization allows each component to be trained with its appropriate objective while sharing a common state encoder.

Formally, we model each episode as a Markov Decision Process $\mathcal{M} = (\mathcal{S}, \mathcal{A}, P, R, \gamma)$,
where $\mathcal{S}$ is the continuous state space encoding (detailed in Section~\ref{ssc:state}); $\mathcal{A} = \mathcal{A}_d \times \mathcal{A}_c$ is the
parameterized hybrid action space; $P(s'|s, a)$ is the transition kernel;
$R(s, a)$ is the scalar reward; and $\gamma \in (0, 1)$ is the discount factor.
To model domain randomization, we adopt a contextual MDP formulation $\mathcal{M}(\omega)$
[see Section~\ref{ss:overestimation}], where the configuration vector $\omega$, encoding object poses,
masses, and friction coefficients $\overset{\mathrm{iid}}{\sim} P(\Omega)$ at the start of each episode and held fixed within it.

\subsubsection{RL baselines for hybrid action spaces}
To identify the most stable algorithmic foundation for hybrid action learning, we evaluated four DRL baselines -- TD3~\cite{fujimoto2018addressing}, SAC~\cite{haarnoja2018soft}
, DDPG~\cite{lillicrap2015continuous}, and PPO~\cite{schulman2017proximal} -- each adapted to the hybrid action space under \emph{full domain randomization}, where object poses, masses, and scene configurations are independently randomized at every episode reset
(Appendix~\ref{sec:hybrid_baselines}). Thus, we obtain four algorithms: HyTD3, HySAC, HyPPO, and HyDDPG, respectively.

As shown in Fig.~\ref{fig:baselines}, HyTD3 consistently achieves superior reward and more stable training curves compared to the other baselines. This empirical advantage is attributable to three structural properties of TD3: (i) its twin-critic architecture, which mitigates overestimation bias via clipped double Q-learning; (ii) delayed policy updates, which reduce the frequency of destructive gradient interference between actor and critic; and (iii) its deterministic policy, which avoids the entropy-regularization instability that hampers SAC under distributional shifts induced by heavy randomization. In contrast, HyDDPG suffers from unmitigated overestimation bias due to its single critic, HySAC struggles to maintain entropy-regularized convergence under non-stationary distributions, and the on-policy nature of HyPPO makes it sample-inefficient under high-variance randomized environments. Motivated by these results, we build our hybrid extension upon the TD3 framework, as detailed in the following subsection.

\begin{figure*}[!ht]
    \centering
    \begin{tikzpicture}[node distance = 0.5cm and 1.1cm]

%% ────────────────────────────────────────────────────────────────────────────
%%  Robot image
%% ────────────────────────────────────────────────────────────────────────────
\node[inner sep=0pt] (robot)
  {\includegraphics[width=3.55cm, trim=12 6 12 6, clip]{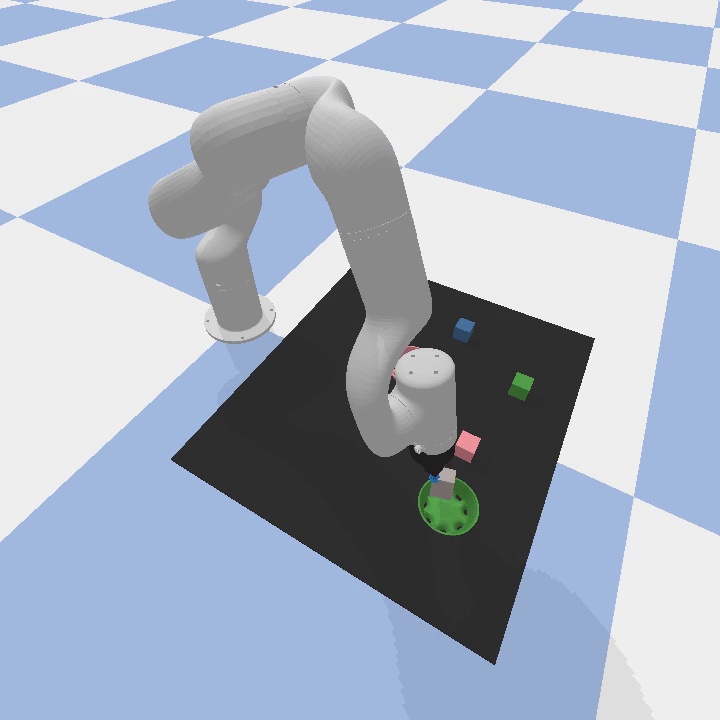}};
\node[font=\small\bfseries, text=Calgo,
      above=0.03cm of robot] {Simulation Environment};

 %% environment & output
\node[envblk,  below=1.4cm of robot,text width=3.5cm]  (env) {Environment};
\node[iobox,   above=0.5cm of env, ,text width=3.5cm]                 (rs)  {$r_t,\;\boldsymbol{s}_{t+1}$};

\node[envblk,  right=1.0cm of robot, ] (state) {State $\boldsymbol{s}_t$};
\node[actor,   right=0.6cm of state]                  (actor) {Actor $\mu_\theta$};

%% discrete / continuous outputs
\node[iobox,   above right=0.3cm and 0.5cm of actor] (ad) {$a^d_t$: mode};
\node[iobox,   below right=0.3cm and 0.5cm of actor] (ac) {$a^c_t$: joint vel.};

%% low-level control chain
\node[intblk,  right=0.4cm of ac]  (integ) {$\int(\cdot)\,dt$};
\node[pdblk,   right=0.4cm of integ] (pd)  {PD Ctrl};

%% — bottom row —
\node[bufblk,  right=1.5cm of rs] (buf){Replay Buffer $\mathcal{D}$};
\node[algoblk, right=1.0cm of buf]                   (algo)   {Hybrid TD3};
\node[critic,  right=2.0cm of algo](critic){Critics $Q_{\phi_1},Q_{\phi_2}$};

% ── Background panel ────────────────────────────────────────────────────────
\begin{scope}[on background layer]
  \node[fill=Cpanel, draw=Cpborder, rounded corners=7pt,
        inner sep=10pt, very thick,
        fit=(state)(actor)(ad)(ac)(integ)(pd)(critic)(buf)(algo)] (panel) {};
\end{scope}
\node[font=\small\bfseries, text=Cpborder!80!black,
      above right=0pt and 0pt of panel.north east,
      anchor=north east, xshift=-5pt, yshift=-3pt]
  {Hybrid TD3 Framework};

\draw[Ag] (robot.east) -- (state.west)
  node[lbl, midway, above] {$\boldsymbol{s}_t$};

%% state → actor
\draw[Ab] (state) -- (actor);

%% actor fork → ad, ac
\draw[Ab] (actor.east) -- ++(0.32,0) |- (ad.west);
\draw[Ab] (actor.east) -- ++(0.32,0) |- (ac.west);

%% continuous chain: ac → integ → pd → env
\draw[A]  (ac)    -- (integ)
  node[lbl, midway, above, xshift=2pt] {};
\draw[A]  (integ) -- (pd)
  node[lbl, midway, above, font=\tiny, yshift=1pt] {$\boldsymbol{j}_t$};
\draw[A]  (pd.east) -- ++(0.6,0) --++(0, -2.7) -- (env.east);

%% discrete: ad → env
\draw[A]  (ad.east)  -- ++(4.635,0) --++(0, -4.685) -- (env.east);

%% env → rs
\draw[Ag] (env) -- (rs);

%% rs → state  (feedback arc, routed above the panel)
\coordinate (feedtop) at ($(rs.east)+(0.3,0)$);
\coordinate (feedup)  at ($(feedtop)+(0,2.4)$);
\draw[Ag]
  (rs.east)
  -- (feedtop)
  -- (feedup)
  -- (state.west)
  node[lbl, pos=-0.06, below, yshift=-68pt] {$\boldsymbol{s}_{t+1}$};

%% rs → buffer  (down then left)
\draw[Ap]
  (rs.east) -- (buf.west);

%% buffer → algo
\draw[Ap] (buf) -- (algo);

%% algo → critics  (below the buffer, left)
\draw[Ao]
  (algo.east) -- (critic.west)
  node[lbl, pos=0.50, above] {update $\phi_1,\phi_2$};

%% critics → actor  (dashed policy gradient, going up then right to actor)
\draw[Ad]
  (critic.north) -- ++(0,0.4)
  -| (actor.south)
  node[lbl, pos=0.5, above, xshift=-8pt, yshift=-2pt] {\rotatebox{90}{\tiny policy gradient}};

%% buffer tuple label
\node[lbl, above=0.0cm of buf] {$(s,\,a^c,\,a^d,\,r,\,s')$};

\end{tikzpicture}
    \caption{Architecture of the proposed Hybrid TD3 framework.}
    % \caption{Our DRL system deviates from the traditional Markov Decision Process (MDP) that not only relies on the current trajectory to decide the future but also combines the past trajectories to help the agent learn smoother. This model processes the environment observation \(o_t\) that consists of the agent's proprioceptive, exteroceptive, relational, and historical data.}
    \label{fig:hybrid_system}
\end{figure*}

\subsection{Hybrid TD3: Weighted Clipped Q-Learning}

We extend the standard TD3 to the hybrid action space by defining twin critics: $Q_{\phi_i}(s, a_c, a_d)$, $i \in \{1, 2\}$, that jointly evaluate both action components. Both critics share the same regression loss:
\begin{equation}
L(\phi) = \mathbb{E}_{(s,a_c,a_d,r,s') \sim \mathcal{D}}
\sum_{i=1}^2 
\left(Q_{\phi_i}(s, a_c, a_d) - y\right)^2,
\end{equation}
where $\mathcal{D}$ is the relay buffer and $y$ is the bootstrapped regression target described below. The standard TD3 target selects the discrete action greedily and applies the clipped minimum over two target critics:
\begin{equation}
    y = r + \gamma \min_{i=1,2}\, (Q_{\phi_i'}(s',\, a_c',\, a_d'),
\end{equation}
where $a_c' = \mu_{\theta'}(s') + \epsilon$, $\epsilon \sim \mathcal{N}(0,\sigma^2)$
is target policy smoothing noise, and $a_d' = \arg\max_{k \in \mathcal{A}_d}
p_d(k \mid s')$ is the greedy discrete action under the target policy.

Instead of committing to a single greedy discrete action, we propose to \emph{marginalize} the clipped minimum target over the full discrete action distribution. The resulting weighted clipped Q-learning target is:
% \begin{equation}
% y = r + \gamma \frac{1}{K} \sum_{k=1}^{K} \pi_d(a_d^{(k)} \mid s') \min_{i=1,2} Q_{\phi_i'}(s',\, a_c',\, a_d^{(k)}).
% \label{eq:critic_loss}
% \end{equation}
\begin{equation}
y = r + \gamma \sum_{k=1}^{K} \pi_d(a_d^{(k)} \mid s') \min_{i=1,2} Q_{\phi_i'}(s',\, a_c',\, a_d^{(k)}).
\label{eq:critic_loss}
\end{equation}
This formulation propagates a soft, distribution-weighted value estimate rather than the hard-max target of standard TD3, reducing variance in gradient signals by propagating a weighted average, an effect most pronounced when the discrete policy $\pi_d$ is diffuse during early training. As established in Theorem~\ref{thm:bias_order}, this modification preserves the same expected bias as the standard TD3 target while improving policy smoothness.

To improve robustness in highly stochastic environments, we retain the clipped double-$Q$ operator in the actor update, deviating from standard TD3, which relies on a single critic. Using a minimum of two critics reduces overestimation bias during policy improvement.
In the parameterized hybrid action setting, the actor objective marginalizes over the discrete action distribution while generating mode-conditioned continuous parameters. The objective is defined as:
\begin{equation}
\mathcal{L}_\theta 
= -\,\mathbb{E}_{s \sim \mathcal{D}}
\left[
\sum_{k=1}^{K}
\pi_d(a_d^{(k)} \mid s)\,
\min_{i=1,2}
Q_{\phi_i}\!\big(s, a_d^{(k)}, a_c\big)
\right].
\label{eq:actor_loss}
\end{equation}
By marginalizing over the discrete policy $\pi_d$ rather than applying a hard $\arg\max$ selection, the objective enables differentiable credit assignment across all discrete modes. This soft aggregation improves optimization stability and accelerates convergence during early training stages, when the discrete policy remains diffuse.

\subsection{Theoretical Analysis of Overestimation Bias}
\label{ss:overestimation}
A central motivation for choosing TD3 as the backbone is its principled handling of overestimation bias. In standard Q-learning~\cite{watkins1992q}, the greedy Bellman target:
% \begin{equation}
$y = r + \gamma \max_{a' \in \mathcal{A}} Q(s', a')$
% \end{equation}
suffers from systematic overestimation because the maximization operator over noisy estimates satisfies $\mathbb{E}[\max_{a'} \hat{Q}(s',a')] \geq \max_{a'} Q^\pi(s', a')$ by Jensen's inequality applied to the convex $\max$ operator. In deep Q-learning, this effect is amplified when the same network is used for both action selection and evaluation, creating a self-reinforcing feedback loop that inflates value estimates across updates~\cite{hasselt2010double}

In hybrid action spaces, overestimation arises from two coupled sources. Within each Bellman update, the agent maximizes over both discrete modes and continuous parameters simultaneously - two maximization operations whose approximation errors combine rather than cancel. This within-step coupling is captured formally in Theorem~\ref{thm:bias_order}, where the bias of the hybrid Bellman target reflects the joint contribution of both action components. Across training updates, inflated value estimates can bias subsequent discrete mode selection, which in turn distorts the continuous optimization landscape - a cross-step interaction that is not formally bounded in this work but is empirically visible in the diverging bias trajectories of HyACC and HyTQC in Fig. \ref{fig:estimation_error}. To rigorously characterize these effects, we derive formal bounds on the expected estimation bias for five hybrid algorithmic variants: Our proposed (Hybrid TD3) and four baseline adaptations—HyACC, HyTQC, HyDARC, and HyDATD3 which represent the extensions of ACC \cite{dorka2022adaptively}, TQC \cite{kuznetsov2020controlling}, DARC \cite{lyu2022efficient}, and DATD3 \cite{lyu2022efficient} to hybrid action spaces, respectively. In addition, We formalize the robotic environment as a parameterized Markov Decision Process (MDP), denoted by $\mathcal{M}(\omega)$, where the configuration vector $\omega \in \Omega$, representing object orientation, location, and target position, is sampled from a distribution $P(\Omega)$ at the onset of each episode. 
% \textbf{Since $\omega$ is randomized per episode, the agent perceives the transition dynamics $\mathcal{P}(s' \mid s, a)$ and reward functions $\mathcal{R}(s, a)$ as non-stationary across the training set. This inherent environmental noise significantly increases the value function's susceptibility to large estimation errors, thereby increasing the likelihood of substantial bias.}
This defines a contextual MDP in which the context $\omega$ is $\overset{\mathrm{iid}}{\sim}$ at episode onset from $P(\Omega)$ and held fixed within the episode.
\begin{theorem}
\label{thm:bias_order}
    Let $Q^{\pi}$ denote the true $Q$ function under the current target policy $\pi$, and let $Q_{\theta}$ be its neural network approximation with estimation error $\epsilon^{k}_{\phi,\xi}$ at stage $k$, where $\xi \in \{1,2\}$ indexes the critic networks. Assume the estimation errors satisfy Assumption 1 (see Appendix~\ref{sec:overestimation}). Let $\delta Q$ denote the expected estimation bias of the Bellman target for each variant, as defined in Appendix~\ref{sec:estimation_algo}. Then the ordering of expected bias, from most negative (strongest underestimation) to least negative, is (\textbf{Proof.} See Appendix~\ref{sec:overestimation}):
    \begin{equation}
    \begin{aligned}
    \delta Q_{\mathrm{\text{Hybrid\_TD3}}}
    &< \delta Q_{\mathrm{HyACC}} \approx \delta Q_{\mathrm{HyTQC}} \\
    &< \delta Q_{\mathrm{HyDARC}} < \delta Q_{\mathrm{HyDATD3}}.
    \end{aligned}
    \label{eq:bias_order}
    \end{equation}
\end{theorem}

\medskip 
Theorem~\ref{thm:bias_order} has two important implications. First, our proposed Hybrid TD3 achieves the lowest bias, confirming that the soft marginalization over the discrete distribution does not introduce additional pessimism. Second, the clipped double Q-learning target produces a tightly bounded, controlled underestimation that is more stable than the distributional methods (HyTQC, HyACC) and less prone to overestimation than HyDATD3. This formally justifies TD3 as the most reliable backbone for hybrid action RL under dense rewards and heavy randomization. 

The causal pathway from domain randomization to elevated estimation bias operates through three steps. First, domain randomization requires the critic to regress a value function over a distribution of $\mathcal{M}(\omega)$ rather than a single stationary environment. A harder regression target that cannot be solved by specializing to a fixed transition kernel. Second, this increased regression complexity tends to induce larger approximation variance $\sigma^2$ in the critic estimates, as the function approximator must simultaneously represent value functions for configurations with different object geometry, mass, and contact properties. Third, under Theorem~\ref{thm:bias_order}, the bias term of the Bellman target scales as $\sigma/\sqrt{\pi}$ so larger $\sigma$ directly amplifies the magnitude of estimation bias in the hybrid setting. This chain is not formally closed: we do not derive $\sigma$ as a function of the randomization range $|\Omega|$. However, Fig.~\ref{fig:estimation_error} provides indirect empirical evidence: the estimation bias of SAC-based methods which lack the variance reduction of clipped double Q-learning diverges substantially under our randomization protocol, while TD3-based methods maintain tightly bounded bias, consistent with the prediction that variance reduction matters more when $\sigma$ is large.

\subsection{Hybrid Action RL Formulation}
\subsubsection{State Representation}
\label{ssc:state}
To enable effective decision making, the agent receives a structured state embedding $\boldsymbol{s}_t$ that captures both robot kinematics, object geometry, and task-relevant relational information. 
A 2-step history of joint position $\boldsymbol{j}_{t-2:t}$ provides implicit velocity information through temporal differences. Relative position vectors $\boldsymbol{p}_{eo}$, $\boldsymbol{p}_{og}$, and $\boldsymbol{p}_{eg}$ encode the geometric relationships between the end-effector, object, and goal, guiding reaching and manipulation while improving invariance to global coordinates.  Suction observations $\boldsymbol{su}_t$ include two binary indicators (activation and release) together with a continuously predicted continuous suction command, enabling joint reasoning over contact state and actuation intent. Previous actions $\boldsymbol{a}_{t-3:t-1}$ are included to promote temporal smoothness and mitigate abrupt control transitions.

All state components are normalized using Welford's algorithm, which maintains running estimates of the mean $\mu_t$ and variance $\sigma_t^2$ :

\begin{equation}
\begin{aligned}
    n_t &= n_{t-1} + 1, \delta_t = \mathbf{x}_t - \mu_{t-1}, \mu_t = \mu_{t-1} + \dfrac{\delta_t}{n_t}, \\
    \delta_t' &= \mathbf{x}_t - \mu_t, M2_t = M2_{t-1} + \delta_t \odot \delta_t'.
\end{aligned}
\end{equation}

The running variance and standard deviation are $\sigma_t^2 = M2_t / n_t$ and $\sigma_t = \sqrt{\sigma_t^2}$. Each observation $\mathbf{x}$ is then standardized as:
$\hat{\mathbf{x}} = 
\dfrac{\mathbf{x} - \mu_t}{\sigma_t + \epsilon}, \qquad \epsilon = 1e-7$,
keeping state elements approximately zero-mean with unit variance throughout training.

\subsubsection{Action Representation}
The continuous action $\boldsymbol{a}_t$ specifies predicted joint velocities $\dot{\boldsymbol{j}}_t$ for all robot joints. Target joint positions are obtained by integrating these velocities over the simulation timestep $dt$:
\begin{equation}
\boldsymbol{j}_t = \boldsymbol{j}_{t-1} + \dot{\boldsymbol{j}}_t \cdot dt.
\label{eq:velocity_integration}
\end{equation}

The resulting joint targets are then tracked by a low-level PD controller that enforces velocity bounds and joint-limit constraints. PD controller introduces history dependence. The next state $s_{t+1}$ depends on the unmodeled internal state of the controller, which can be approximated as a dependence on recent actions. We address this by
augmenting the state with the previous actions $a_{t-2:t-1}$, recovering an approximately Markovian input representation.

%%%%%%%%%%%%%%%%%%%%%%%%%%%%%%%%%%%%%%%%%%%%%%%%%%%%%%%%%%%%%%%%%%%%%%%%%%%%%%%%%%%%%%%%%%%%%%%%%%%%%
%%%%%%%%%%%%%%%%%%%%%%%%%%%%%%%%%%%%%%%%%%%%%%%%%%%%%%%%%%%%%%%%%%%%%%%%%%%%%%%%%%%%%%%%%%%%%%%%%%%%%
\subsubsection{Reward \& Penalty Design}
We adopt the hierarchical reward and penalty formulation from previous
work~\cite{canh2026human} and summarize it here for completeness. The total reward at each timestep is a weighted combination of three reward terms and six penalty terms:
\begin{equation}
R_t(\boldsymbol{s}_t, \boldsymbol{a}_t)
= \sum_{i=0}^{2} W_i \cdot r_i^t(\boldsymbol{s}_t, \boldsymbol{a}_t)
- \sum_{j=0}^{5} W_j \cdot c_j^t(\boldsymbol{s}_t, \boldsymbol{a}_t),
\label{eq:total_reward}
\end{equation}
normalized by a constant factor of $10$ to stabilize critic gradient magnitudes.

The reward terms provide stage-dependent supervision: $r_{0,t}$ encourages safe end-effector approach to the target object; $r_{1,t}$ measures goal-directed transport progress conditioned on established object contact; and $r_{2,t}$ rewards object velocity aligned with the goal direction. The penalty terms discourage collisions with the ground, the robot itself, or surrounding objects; penalize exceeding the maximum episode length; and constrain end-effector inclination, action jerkiness, object tilt, and workspace boundary violations.

\subsubsection{Training Protocol}

The overall DRL framework is illustrated in Fig.~\ref{fig:hybrid_system}. The agent is trained using Hybrid TD3 across four manipulation tasks: \textit{reach}, \textit{pick}, \textit{move}, and \textit{put}. At the start of each episode, the simulator randomly instantiates a scene containing more than 21 distinct object categories, with randomized poses, masses, and friction coefficients. For the reaching task, the agent must bring the end-effector to a target object. For the remaining tasks, success requires completing two sequential sub-goals: correctly approaching the target object and transporting it to the goal location. We adopt a sequential episode-level randomization protocol: at each episode reset, a single configuration $\omega \sim P(\Omega)$ is drawn independently and held fixed for the duration of that episode. In contrast to simultaneous multi-environment training, where N parallel instances maintain instantaneous buffer diversity, sequential randomization produces a replay buffer whose empirical distribution converges to $P(\Omega)$ only as training accumulates episodes. During the early training phase, this constitutes a stricter early-phase generalization test under limited replay diversity, as the critic is updated on a less varied mixture of configurations before sufficient episodes have accumulated. We adopt this protocol as it reflects practical constraints where parallel simulation infrastructure is unavailable, and note that the consistent convergence of Hybrid TD3 under this setting is evidence of robustness to the early-phase diversity deficit.

An episode terminates when the agent successfully completes the required task, or upon any of the following failure conditions: end-effector collision with the ground or self-collision, a joint angle exceeding its allowable range, the object leaving the workspace, or the episode length exceeding a maximum of $100$ steps. This termination scheme encourages the agent to develop safe, efficient trajectories while penalizing unstable behaviors.

%===============================================================================

%===============================================================================
\section{Experimental Results}
\label{sec:result}
\begin{figure}[htbp]
\centering
\begin{tabular}{cc}
\includegraphics[width=0.22\textwidth]{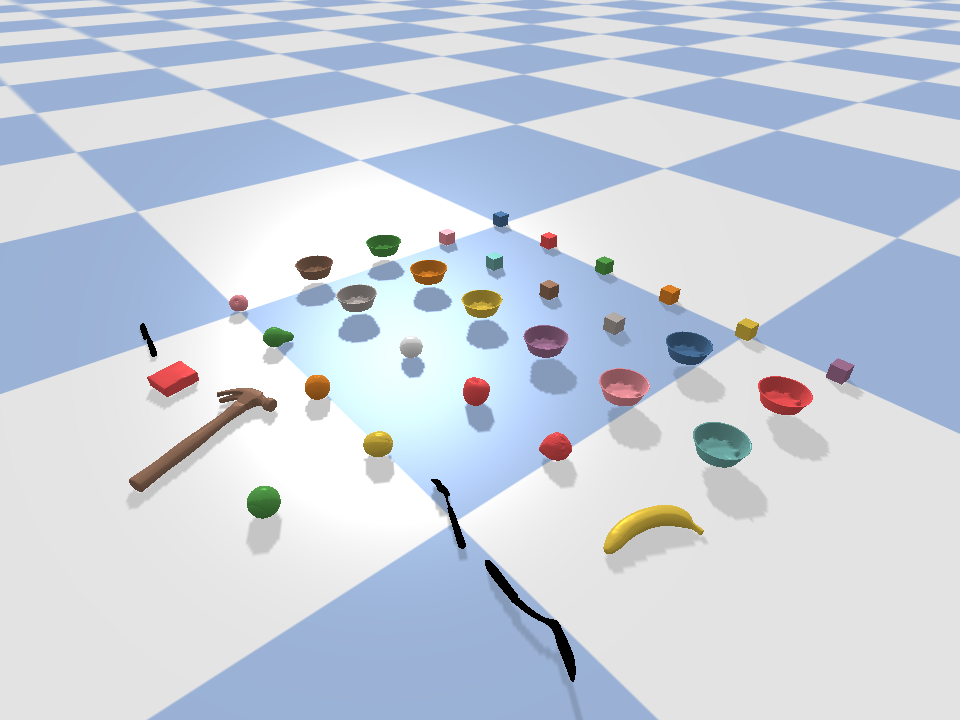} &
\includegraphics[width=0.22\textwidth]{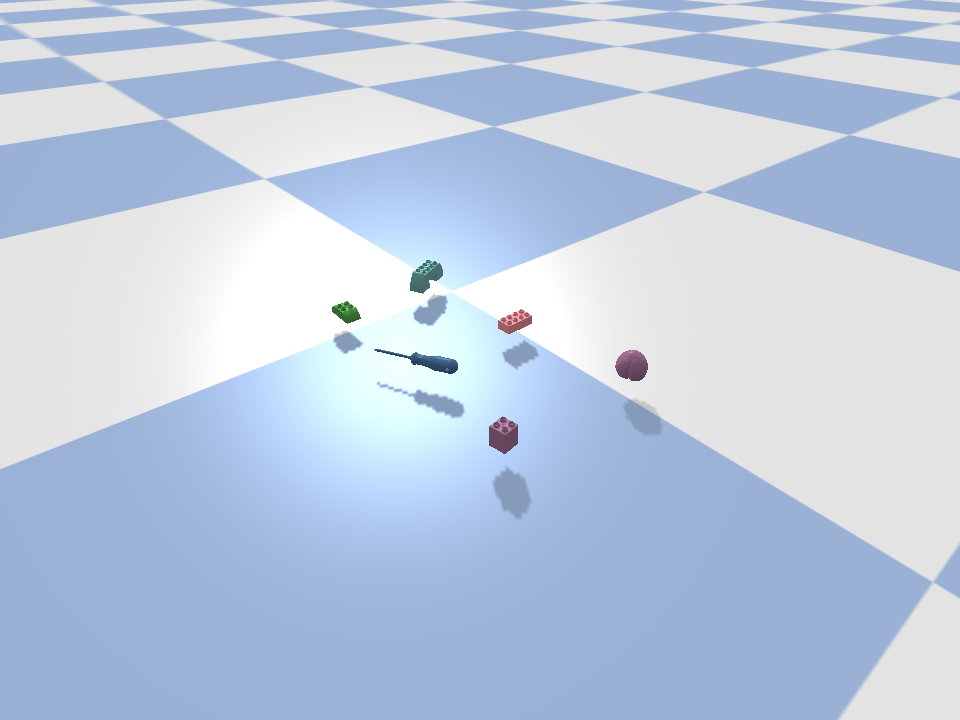}
\end{tabular}
\caption{Training (left) and test objects (right) used for zero-shot generalization evaluation}
\label{fig:two_images}
\end{figure}

\begin{figure*}[!ht]
    \centering
    \includegraphics[width=\textwidth]{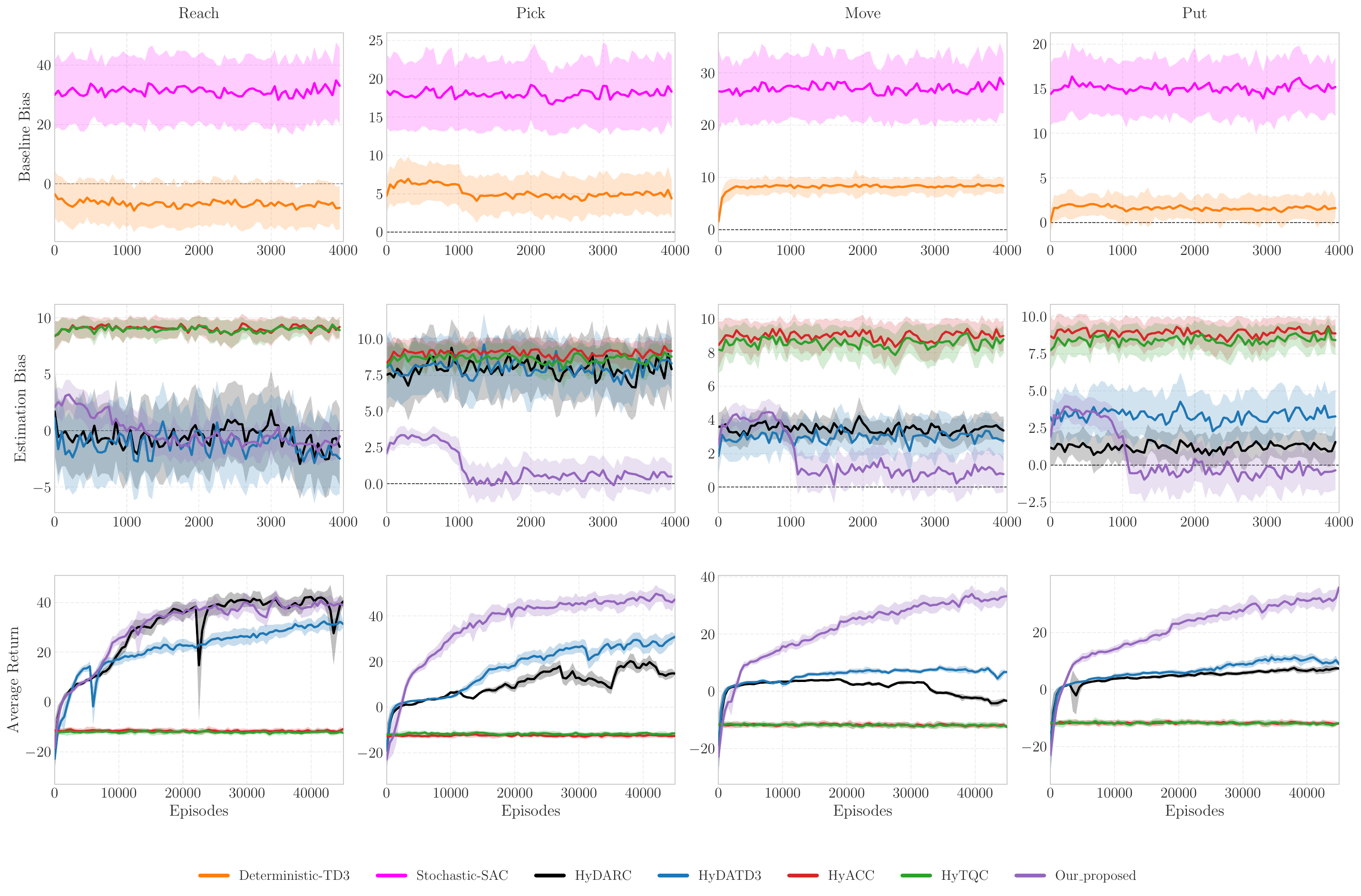}
    \caption{Estimation bias of the baselines (top row), estimation bias of the proposed methods (middle row), and average return (bottom row) across four manipulation tasks. Solid curves represent mean performance, while shaded areas indicate standard deviations over four independent random seeds.}
    \label{fig:estimation_error}
\end{figure*}

\begin{figure*}[!ht]
    \centering
    \includegraphics[width=\textwidth]{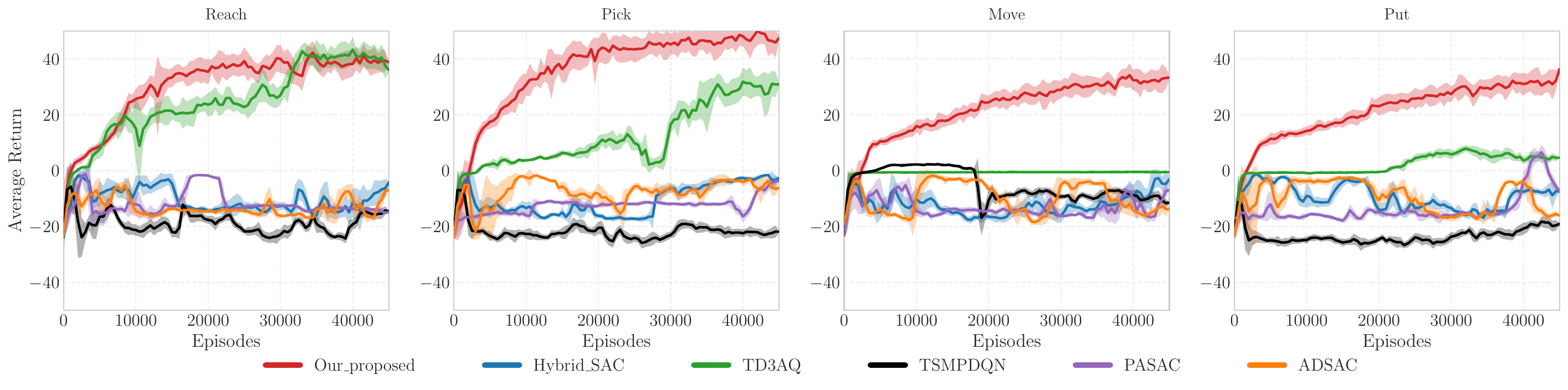}
    \caption{Average return learning curves across four manipulation tasks.}
    \label{fig:reward}
\end{figure*}

\begin{figure*}[!ht]
    \centering
    \includegraphics[width=\textwidth]{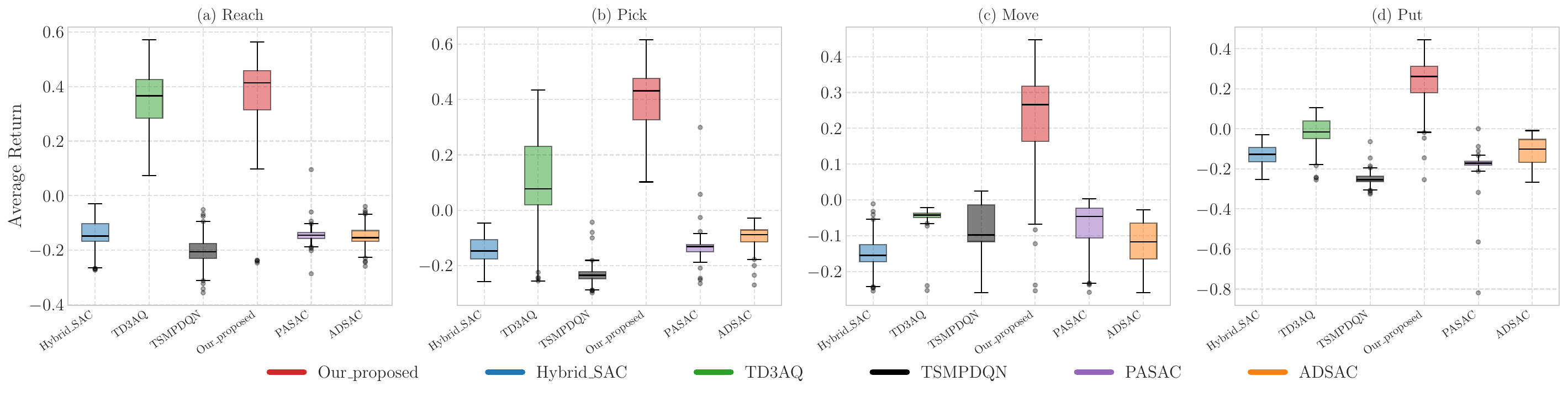}
    \caption{Distribution of final episode rewards across seeds and evaluation episodes for our proposed and hybrid action baselines}
    \label{fig:boxplot}
\end{figure*}

\begin{figure*}[!ht]
    \centering
    \includegraphics[width=\textwidth]{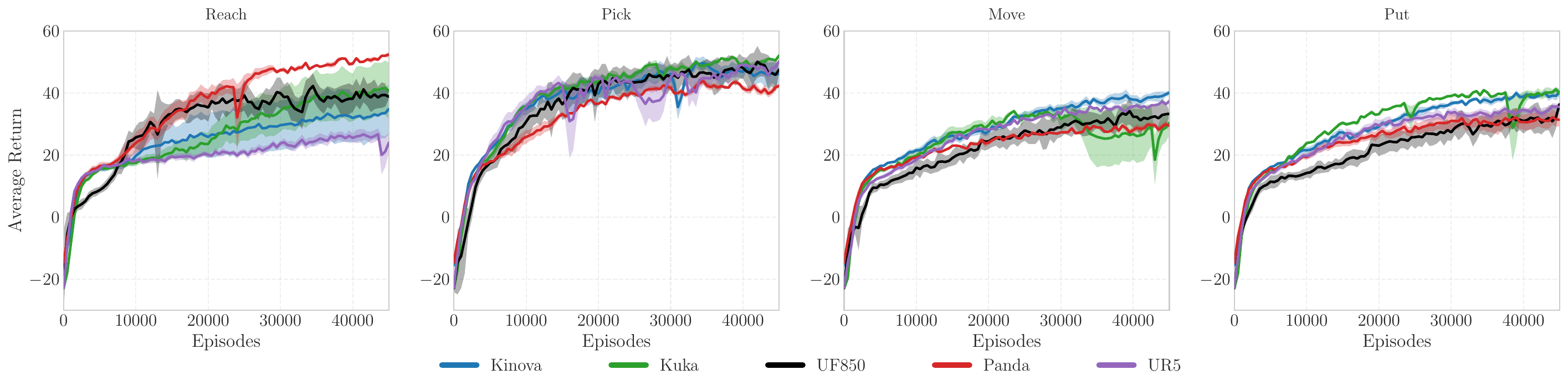}
    \caption{Average return learning curves across four manipulation tasks for 5 different types of robotic arm.}
    \label{fig:reward_robot}
\end{figure*}

\begin{figure*}[!ht]
    \centering
    \includegraphics[width=\textwidth]{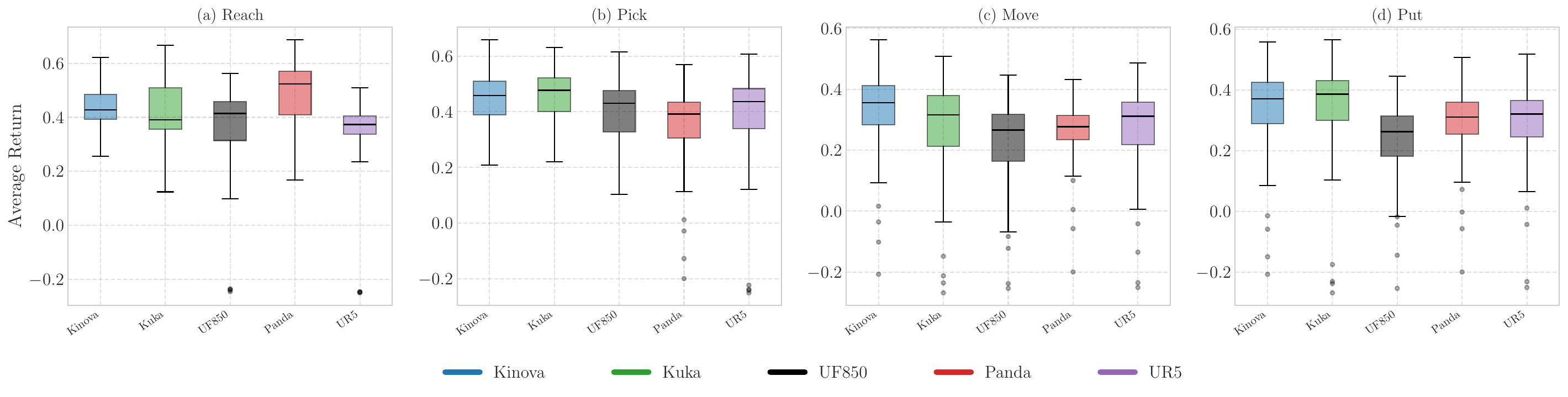}
    \caption{Distribution of final episode rewards across seeds and evaluation episodes for 5 different types of robotic arm}
    \label{fig:boxplot_robot}
\end{figure*}

\begin{figure*}[!ht]
    \centering
    \includegraphics[width=0.9\textwidth]{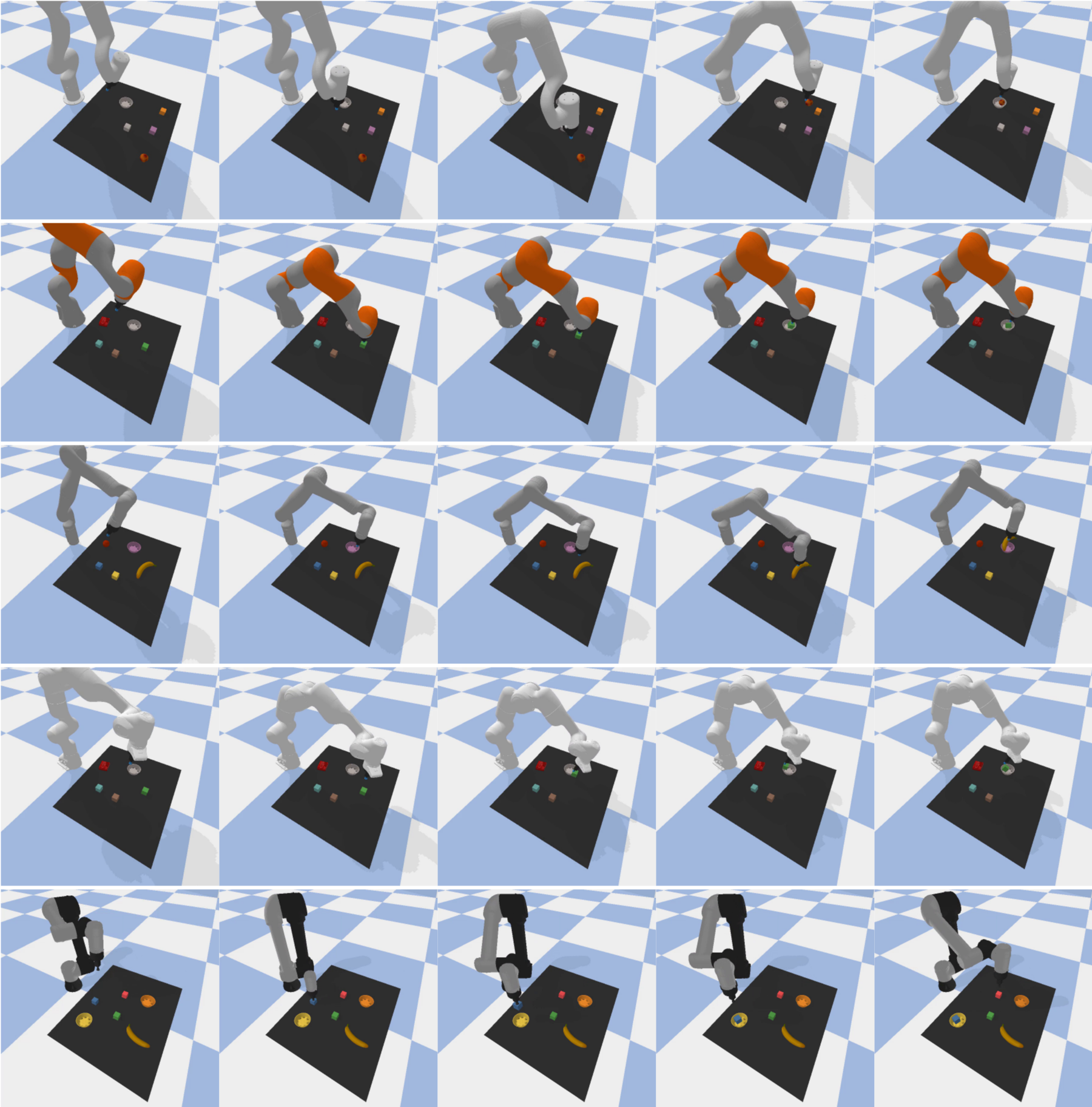}
    \caption{Representative execution trajectories of 5 different types of robotic arm}
    \label{fig:qualitative}
\end{figure*}
% This section evaluates Hybrid TD3 across three dimensions: (i) comparison against state-of-the-art hybrid action RL baselines on four manipulation tasks; (ii) analysis of estimation bias versus average return to validate the theoretical bias ordering derived in Theorem~\ref{thm:bias_order}; and (iii) .
%-------------------------------------------------------------------------------
\subsection{Experimental Setup}
\label{sec:exp_setup}

\subsubsection{Simulation Environment}
All experiments are conducted in PyBullet~\cite{coumans2016pybullet} using a UF850 robotic arm equipped with a suction gripper. The agent operates in joint space, outputting 6-DOF joint velocities as the continuous action component and a binary suction mode as the discrete action component. To ensure the generalization capability enabled by full domain randomization, we test the trained policy on a set of novel object categories that were \emph{never} encountered during training. As shown in Fig.~\ref{fig:two_images}, the test objects differ substantially from the training set in shape, size, color, and surface texture, constituting a meaningful out-of-distribution evaluation. 

\subsubsection{Baselines}
We compare Hybrid TD3 against two groups of baselines. The first group consists of methods for mitigating overestimation bias in hybrid action spaces: HyDARC~\cite{lyu2022efficient}, HyDATD3~\cite{lyu2022efficient}, HyACC~\cite{dorka2022adaptively}, and HyTQC~\cite{kuznetsov2020controlling}. The second group consists of dedicated hybrid action RL algorithms from the literature: Hybrid-SAC \cite{delalleau2019discrete}, TD3AQ \cite{xu2023mixed}, TSMPDQN \cite{zhang2022learning}, PASAC \cite{lin2024discretionary}, and ADSAC \cite{xu2023action}. All baselines are adapted to the same parameterized hybrid action space (binary discrete mode, 6-DOF continuous joint velocities) and trained under identical domain randomization conditions for a fair comparison.
% \textcolor{red}{Since no standardized hybrid-action implementations exist for these bias-mitigation algorithms, we extend each method to the parameterized hybrid setting using a consistent adaptation principle: the critic objective is augmented to accept both action components as input, and the actor objective is extended via discrete marginalization (Eq. 5), consistent with our proposed framework. For TD3-based methods (HyDARC, HyDATD3), the greedy discrete selection is replaced by the marginalization target defined in Appendix~\ref{sec:estimation_algo} for SAC-based methods (HyACC, HyTQC), the entropy regularization is applied independently to both action components with separate temperature parameters, following the adaptation in Appendix~\ref{sec:estimation_algo}. This ensures that performance differences reflect algorithm-level design choices, specifically the critic target construction rather than implementation-level artifacts. All baselines use the same hyperparameter configuration as Hybrid TD3 (learning rate, batch size, buffer size, network architecture); no per-algorithm tuning was performed.}
Since no standardized hybrid-action implementations exist for these bias-mitigation baselines, we extend each method to the parameterized hybrid setting using a consistent adaptation principle detailed in Appendix~\ref{sec:estimation_algo}. This ensures that performance differences reflect algorithm-level design choices, specifically critic target construction rather than implementation artifacts. All baselines share identical hyperparameters with Hybrid TD3; no per-algorithm tuning was performed.

\subsubsection{Training Details} Each agent is trained for $45{,}000$ episodes with a maximum of $100$ steps per episode. Performance is evaluated every epoch using the undiscounted cumulative reward over a held-out test episode. All results are reported as the mean and standard deviation across $4$ independent random seeds. The estimation bias is computed as the mean difference between the Monte Carlo return and the predicted Q-value for each state-action pair visited during test episodes.

%-------------------------------------------------------------------------------
\begin{table*}[!ht]
\centering
\caption{Task success rates (\%) over 100 trials per action, averaged across four random seeds (mean $\pm$ std). 
\colorbox{best}{\phantom{xx}}
and
\colorbox{second}{\phantom{xx}}
indicate the best and second-best results, respectively.}
\label{tab:success_rate}
\begin{tabularx}{\textwidth}{
    >{\raggedright\arraybackslash}X 
    >{\centering\arraybackslash}p{0.18\textwidth} 
    >{\centering\arraybackslash}p{0.18\textwidth} 
    >{\centering\arraybackslash}p{0.18\textwidth} 
    >{\centering\arraybackslash}p{0.18\textwidth}
}
\toprule
\textbf{Robot Platform} & \textbf{Action 0} & \textbf{Action 1} & \textbf{Action 2} & \textbf{Action 3} \\
\midrule
\multicolumn{5}{l}{\textit{Standard Environment (Normal Setting)}} \\
\midrule
UF850       & $94.25 \pm 1.92$ & $89.75 \pm 4.66$ & $80.75 \pm 2.58$ & $83.25 \pm 3.56$ \\
KUKA        & \textbf{\colorbox{best}{$96.67 \pm 1.25$}} & \textbf{\colorbox{second}{$90.00 \pm 2.94$}} & \textbf{\colorbox{second}{$83.00 \pm 4.08$}} & \textbf{\colorbox{best}{$90.67 \pm 2.05$}} \\
Panda       & \textbf{\colorbox{second}{$94.67 \pm 2.05$}} & $74.67 \pm 2.05$ & $69.00 \pm 1.63$ & $83.33 \pm 2.05$ \\
Kinova      & \textbf{\colorbox{second}{$94.67 \pm 2.05$}} & \textbf{\colorbox{best}{$91.33 \pm 2.87$}} & \textbf{\colorbox{best}{$89.33 \pm 1.70$}} & \textbf{\colorbox{second}{$87.33 \pm 1.70$}} \\
UR5e        & $88.00 \pm 2.50$ & $82.00 \pm 3.00$ & $75.00 \pm 4.50$ & $84.00 \pm 2.00$ \\
\midrule
\multicolumn{5}{l}{\textit{Unseen Environment (Generalization Setting)}} \\
\midrule
UF850       & \textbf{\colorbox{second}{$94.25 \pm 1.92$}} & \textbf{\colorbox{second}{$90.00 \pm 5.15$}} & $81.75 \pm 4.66$ & $82.75 \pm 2.58$ \\
KUKA        & $94.00 \pm 2.16$ & $89.67 \pm 1.89$ & \textbf{\colorbox{second}{$82.33 \pm 5.44$}} & \textbf{\colorbox{second}{$86.00 \pm 4.32$}} \\
Panda       & $89.33 \pm 6.02$ & $72.00 \pm 2.45$ & $62.67 \pm 1.89$ & $81.67 \pm 0.94$ \\
Kinova      & \textbf{\colorbox{best}{$94.67 \pm 2.05$}} & \textbf{\colorbox{best}{$91.33 \pm 2.87$}} & \textbf{\colorbox{best}{$89.33 \pm 1.70$}} & \textbf{\colorbox{best}{$87.33 \pm 1.70$}} \\
UR5e        & $84.00 \pm 3.20$ & $78.00 \pm 3.00$ & $70.00 \pm 5.20$ & $80.00 \pm 2.00$ \\
\bottomrule
\end{tabularx}
\end{table*}

\subsection{Comparison Against Bias Mitigation Variants}
\label{sec:bias_results}

Fig.~\ref{fig:estimation_error} presents the estimation bias (top, middle rows) and average return (bottom row) for Hybrid TD3 and the four bias-mitigation baselines across all four tasks. These results provide direct empirical validation of the theoretical ordering established in Theorem~\ref{thm:bias_order}.

\paragraph{Estimation bias} 

The top row of Fig.~\ref{fig:estimation_error} reveals that HyACC and HyTQC exhibit substantially higher positive bias than all the variants - a consequence of their SAC-based, which introduces stochastic entropy-regularized update. As shown by the Stochastic-SAC baseline, which exhibits strong overestimation bias under full-domain randomization, this bias propagates directly into the HyACC and HyTQC targets despite their quantile-truncation correction. In contrast, deterministic TD3-based methods - our proposed Hybrid TD3 maintains tightly bounded, near-zero or small negative bias, and HyDARC, HyDATD3 in most tasks (because in action 1 column has indicated that these two methods have large overestimation bias below HyACC and HyTQC), consistent with the theoretical ordering. Although Eq.~\eqref{eq:bias_order} places HyACC and HyTQC between the TD3-based variant, the SAC-induced positive bias effectively shifts HyACC and HyTQC into the overestimation regime, placing them above the TD3-based group empirically. This highlights a practical scope of the theoretical ordering: Theorem~\ref{thm:bias_order} characterizes the bias of the Bellman target in isolation, whereas the observed estimation bias additionally integrates actor approximation error — which for stochastic SAC-based actors grows substantially under aggressive domain randomization. This discrepancy is itself informative: actor stochasticity constitutes an independent source of overestimation under distributional shift, one that deterministic actor structure suppresses and that the theorem does not claim to bound. The theoretical and empirical results thus provide complementary evidence for the superiority of deterministic TD3-based architectures in this regime. For the Reach task, Deterministic-TD3, HyDARC, HyDATD3, and our proposed method all exhibit underestimation (negative bias). The Reach task is comparatively simple, it requires only an end-effector approach without contact establishment so critics converge reliably, and the clipped double Q-learning target produces stable, tightly bounded underestimation as expected.

\paragraph{Average return} Hybrid TD3 achieves the highest final return across all four tasks, with the gap most pronounced in Pick and Put where contact-rich precision amplifies the consequences of overestimation. HyDATD3, despite the least negative bias, shows high training variance, confirming that moving bias toward overestimation is detrimental under heavy randomization. HyACC and HyTQC converge more slowly with greater variance, while HyDARC is competitive on Reach but degrades on multi-stage tasks.
%-------------------------------------------------------------------------------
\subsection{Comparison Against Hybrid Action Baselines}
\label{sec:hybrid_results} 

Fig.~\ref{fig:reward} and Fig.~\ref{fig:boxplot} compare Hybrid TD3 against dedicated hybrid action algorithms. Hybrid TD3 achieves faster convergence, higher asymptotic reward, and the tightest final performance distribution (highest median, smallest IQR) across all four tasks. TSMPDQN and PASAC exhibit volatile early training with frequent reward collapses in Pick and Move, while Hybrid-SAC and ADSAC are more stable but converge to substantially lower rewards. TD3AQ is competitive on Reach but degrades on multi-stage tasks, where its discrete-action approximation struggles with the parameterized pick-and-place structure. The tight reward distributions of Hybrid TD3 in Fig.~\ref{fig:boxplot} contrast sharply with the wide spreads of TSMPDQN and PASAC, confirming that the weighted clipped Q-learning target produces more reliable policies under randomized evaluation.

\subsection{Quantitative and Qualitative results}
Table~\ref{tab:success_rate} reports success rates over 100 trials per task across 4 seeds for both training and unseen object sets. The policy achieves high performance on the standard set ($94.25$\%, $89.75$\%, $80.75$\%, $83.25$\% for Reach through Put) and transfers with minimal degradation to the unseen set, confirming that full domain randomization enables genuine zero-shot generalization without task-specific adaptation. 
To assess whether the training stability of Hybrid TD3 is contingent on the kinematic structure of the primary test platform (UF850), we train independent policies for 4 additional robot arms (KUKA, Panda, Kinova, and UR5e) using identical hyperparameters, network architecture, and reward formulation, with no platform-specific modifications. These platforms differ substantially in joint count ($6-DOF$ vs. $7-DOF$), workspace volume, link inertia, and end-effector geometry. Table~\ref{tab:success_rate} shows that Hybrid TD3 achieves consistent performance across platforms with substantially different joint-space geometry without platform-specific hyperparameter tuning. This provides empirical evidence that the bias control mechanism of Theorem~\ref{thm:bias_order} is not sensitive to the specific structure of the continuous action manifold. 
Fig.~\ref{fig:reward_robot} shows that all five platforms converge stably across all four tasks without reward collapse, confirming that training dynamics remain well-behaved regardless of kinematic structure. Fig.~\ref{fig:boxplot_robot} further shows that final reward distributions remain concentrated across seeds, with Panda exhibiting slightly wider variance on contact-rich tasks. In addition, Fig.~\ref{fig:qualitative} illustrates representative trajectories across all four tasks, showing consistent end-effector positioning across varied randomized scene configurations. Success rate across 100 evaluation trials is reported for Hybrid TD3 only, as the primary baselines (Hybrid-SAC, TD3AQ, TSMPDQN, PASAC, ADSAC) exhibit training instability, particularly reward collapse in Pick and Move tasks visible in Fig.~\ref{fig:estimation_error} that prevents reliable success-rate measurement

%===============================================================================

\section{Conclusion}
\label{sec:conclusion}
This paper presented Hybrid TD3, a principled extension of TD3 to discrete-continuous hybrid action spaces for robotic manipulation under full domain randomization. We established a formal bias ordering across five hybrid algorithmic variants under the Gaussian synchronized error assumption, proving that the weighted clipped Q-learning target, which marginalizes over the discrete action distribution, achieves the lowest expected estimation bias among the variants analyzed. Our analysis further characterized why tightly bounded negative bias is actively beneficial under the non-stationary distributions induced by aggressive randomization, where methods that reduce this bias introduce additional variance and instability. Experiments on four manipulation tasks confirmed theoretical bias ordering and demonstrated superior performance over both dedicated hybrid action baselines and bias-mitigation variants. The policy also generalized zero-shot to novel object categories and was successfully transferred to an alternative robot platform without retraining. Finally, all experiments are conducted in simulation; bridging to physical hardware involves contact dynamics and sensor noise that our analysis does not capture. Future work will focus on relaxing the independence assumption to account for correlated critic errors, deriving formal bounds on approximation variance as a function of randomization range, and validating the approach on physical robot platforms.

%===============================================================================

%===============================================================================
\appendices

\section{RL Baselines for Hybrid Action Spaces}
\label{sec:hybrid_baselines}

All four baselines are adapted to the parameterized \emph{hybrid action space} $(a_c \in \mathcal{A}_c,\; a_d \in \mathcal{A}_d)$ by extending their critic and actor objectives as summarized in Table~\ref{tab:baselines}. For TD3 and DDPG, the actor selects $a_d$ greedily; for SAC, both components are stochastic with separate entropy coefficients $\alpha_c, \alpha_d$ tuned via dual gradient descent; for PPO, the policy factorizes into independent categorical and Gaussian heads with a shared importance-sampling ratio $r_t(\theta) =\dfrac{p_c^{\mathrm{new}}(a_c \mid s)\, p_d^{\mathrm{new}}(a_d \mid s)}{p_c^{\mathrm{old}}(a_c \mid s)\, p_d^{\mathrm{old}}(a_d \mid s)}$.

\begin{table*}[h]
\centering
\caption{Hybrid adaptation of RL baselines. All share the same state
$(s)$ and action $(a_c, a_d)$ inputs to the critics.}
\label{tab:baselines}
\renewcommand{\arraystretch}{1.25}
\begin{tabular}{@{}llll@{}}
\toprule
\textbf{Algorithm} & \textbf{Critic loss} $\mathcal{L}_\phi$& \textbf{Critic target $y$} & \textbf{Actor loss $\mathcal{L}_\theta$} \\
\midrule

TD3  
  & $\sum_{i=1}^{2}\big(Q_{\phi_i}(s, a_c, a_d) - y\big)^2$
  & $r+\gamma\min_{i=1,2}Q_{\phi_i'}(s',a_c',a_d')$ 
  & $-\max_{a_d \in \mathcal{A}_d}Q_{\phi_1}(s,\mu_\theta(s),a_d)$ \\
DDPG 
  & $\big(Q_\phi(s, a_c, a_d) - y\big)^2$
  & $r+\gamma Q_{\phi'}(s',a_c',a_d')$ 
  & $-\max_{a_d \in \mathcal{A}_d}Q_\phi(s,\mu_\theta(s),a_d)$ \\
SAC  
  & $\sum_{i=1}^{2}\big(Q_{\phi_i}(s, a_c, a_d) - y\big)^2$
  & $\begin{aligned}
      r+\gamma\big(\min_{i=1,2} Q_{\phi_i'}&-\alpha_c\log p_c(a'_c|s') \\ &- \alpha_d\log p_d(a'_d|s')\big)
  \end{aligned}$
  & $\begin{aligned}
      \mathbb{E}_{a_c \sim p_c,\, a_d \sim p_d}[\alpha_c\log p_c(a'_c|s')+\alpha_d\log p_d(a'_d|s') - \min_{i=1,2} Q_{\phi_i}]
  \end{aligned}$ \\
PPO  
  & $(V(s) - V_\mathrm{targ})^2$
  & $V(s)+A(s,a)$ (GAE) 
  & $-\mathbb{E}\bigg[\min\big(r_t(\theta) A(s,a),\;
\text{clip}(r_t(\theta), 1-\epsilon, 1+\epsilon) A(s,a) \big )\bigg]$ \\
\bottomrule
\end{tabular}
\end{table*}

%===============================================================================
\section{Estimation Bias Variants for Hybrid Action Spaces}
\label{sec:estimation_algo}

All five variants share the twin-critic squared Bellman loss
\begin{equation}
\mathcal{L}_\phi = \sum_{i=1}^{2}\bigl(Q_{\phi_i}(s,a_c,a_d)-y\bigr)^2,
\label{eq:hybrid_td3_critic_loss}
\end{equation}
and differ only in how the Bellman target $y$ is constructed.
The actor loss for all TD3-based variants marginalizes over the discrete
policy:
$\mathcal{L}_\theta = -\sum_{k} \pi_d(k|s)\min_{i}Q_{\phi_i}(s,\mu_\theta(s),k)$.

\noindent
\textbf{Hybrid TD3.} All hybrid TD3 variants employ two critic networks 
$Q_{\phi_i}(s, a_d, a_c)$, $i \in \{1,2\}$, which are trained using a shared squared Bellman error objective. The critic loss follows Eq.~\eqref{eq:hybrid_td3_critic_loss}. The actor update deviates from standard TD3 and is defined as in Eq.~\ref{eq:actor_loss}.

\noindent
\textbf{HyDATD3~\cite{lyu2022efficient}.} Two independent continuous samples $a_{c1}',a_{c2}'$ from the target actor;
the target takes the more optimistic aggregation:
\begin{equation}
y=r+\gamma\max\!\bigl(\mathcal{T}(a_{c1}'),\mathcal{T}(a_{c2}')\bigr),
\end{equation}
with $\mathcal{T}(a_c')=\sum_k p_d(k|s')\min_i Q_{\phi_i'}(s',a_c',k)$. Letting $S_j=Q^\pi+\eta_j$, $\eta_j=\sum_k p_d(k)\min(\epsilon_{\phi_1}^{j,k},\epsilon_{\phi_2}^{j,k})$. The actor is updated in an alternating fashion, using the corresponding critic:
\begin{equation}
    \mathcal{L}_\theta^{(i=1,2)} =-\sum_{k=1}^{K} p_d(k \mid s)\, Q_{\phi_{i=1,2}}(s,a_c,k).
\end{equation}

\noindent
\textbf{HyDARC~\cite{lyu2022efficient}.} Convex combination of pessimistic and optimistic targets ($\lambda\in[0,1]$):
\begin{equation}
y=r+\gamma\bigl(\lambda\min(S_1,S_2)+(1-\lambda)\max(S_1,S_2)\bigr).
\end{equation}

\noindent
\textbf{HyTQC~\cite{kuznetsov2020controlling}.} Ensemble of $M$ critics each producing $N$ quantile atoms; keeps the lowest $kN$. The critic loss is defined as
\begin{equation}
\mathcal{L}_\phi
=\frac{1}{kMN}
\sum_{m=1}^{M}
\sum_{i=1}^{kN}
\bigl(
Q_m^{(i)}(s,a_c,a_d) - y
\bigr)^2,
\end{equation}
where the target is obtained by truncating the top quantiles:
\begin{equation}
y=r+\gamma\frac{1}{kN}\sum_{i=1}^{kN}Q_{(i)}(s',a_c',a_d').
\end{equation}
The actor objective incorporates entropy regularization for both action types:
\begin{equation}
\begin{aligned}
    \mathcal{L}_\theta
=\mathbb{E}_{a_c,a_d}\bigg[\alpha_c \log p_c(a_c \mid s)+\alpha_d \log p_d(a_d \mid s) \\ 
-\frac{1}{MN} \sum_{m=1}^{M} \sum_{i=1}^{N} Q_m^{(i)}(s,a_c,a_d) \bigg].
\end{aligned}
\end{equation}

\noindent
\textbf{HyACC~\cite{dorka2022adaptively}.} Further removes $\beta$ atoms beyond HyTQC truncation:
\begin{equation}
y=r+\gamma\frac{1}{kN-\beta}\sum_{i=1}^{kN-\beta}Q_{(i)}(s',a_c',a_d').
\end{equation}

%===============================================================================
\section{Estimation Analysis and Proof of Theorem~\ref{thm:bias_order}}
\label{sec:overestimation}

% \noindent\textbf{Assumption 1 (Synchronized Bias Shift).}\label{ass:sync_bias}
% The estimation error $\epsilon^k_{\phi,\xi}$ of each critic $\xi\in\{1,2\}$
% is independent of $(s_k,a_k)$ and has finite mean $\mu_\xi<\infty$. Given that the critic networks are architecturally identical, initialized from the same distribution, and optimized over a common data manifold using symmetric update rules. Since all critics share the same replay buffer, architecture, and update rules, they exhibit a synchronized bias: $\mu_1\approx\mu_2\approx\mu$.

\noindent\textbf{Assumption 1 (Gaussian Error with Synchronized Statistics).}\label{ass:sync_bias}  The estimation error $\epsilon^k_{\phi,\xi}$ of critic $\xi\in\{1,2\}$ at stage k is modeled as approximately Gaussian with mean $\mu_\xi$ and variance $\sigma^2$, and treated as independent across critics for analytical tractability. Since both critics share identical architecture, initialization distribution, replay buffer, and symmetric update rules, their error statistics are synchronized: $\mu_1\approx\mu_2\approx\mu$ and both exhibit the same variance $\sigma^2$. In practice, critics trained on shared replay data may exhibit non-zero cross-critic correlation; we treat independence as a standard theoretical idealization

\noindent\textbf{Lemma 1 (Gaussian Min-Max).}\label{lem:expectation_minmax}
For independent $X\sim\mathcal{N}(\mu_1,\sigma^2)$, $Y\sim\mathcal{N}(\mu_2,\sigma^2)$, evaluating $\mathbb{E}[|X-Y|]$ via its Gaussian integral:
\begin{equation}
\begin{aligned}
    \mathbb{E}[\min(X,Y)]
= \frac{\mu_1 + \mu_2}{2}
- \frac{\sigma}{\sqrt{\pi}}
\exp\!\left(-\frac{(\mu_1 - \mu_2)^2}{4\sigma^2}\right) \\
- \frac{\mu_1 - \mu_2}{2}
\mathrm{erf} \left(\frac{\mu_1 - \mu_2}{2\sigma}\right).
\end{aligned}
\end{equation}
Then applying Assumption~1 ($\mu_1\approx\mu_2\approx\mu$) yields:
\begin{equation}
\mathbb{E}[\min(X,Y)]\approx\mu-\tfrac{\sigma}{\sqrt{\pi}},
\mathbb{E}[\max(X,Y)]\approx\mu+\tfrac{\sigma}{\sqrt{\pi}}.
\label{eq:min_expectation}
\end{equation}

\noindent\textbf{Lemma 2 (Approximation of Nested Min-Max).}\label{lem:nested_operator}
For $W_1=\min(X,Y)$ and $W_2=\min(E,F)$ i.i.d.\ with mean $\mu_w = \mathbb{E}[min(X,Y)]$ and
std $\sigma_w = = \mathrm{Var}[\min(X,Y)]
= \mathbb{E}[\min(X,Y)^2] - \mu_w^2$, applying Lemma~1 to the pair $(W_1,W_2)$ under
Assumption~1 can be approximated by:
\begin{equation}
\begin{aligned}
    \mathbb{E}[\min(W_1,W_2)]\approx\mu-\tfrac{\sigma}{\sqrt{\pi}}-\tfrac{\sigma_w}{\sqrt{\pi}},\\
\mathbb{E}[\max(W_1,W_2)]\approx\mu-\tfrac{\sigma}{\sqrt{\pi}}+\tfrac{\sigma_w}{\sqrt{\pi}}.
\end{aligned}
\label{eq:min_nested}
\end{equation}

\noindent\textbf{Proof of Theorem~\ref{thm:bias_order}.}
Write $Q_{\phi_i}=Q^\pi+\epsilon_{\phi_i}$ with independent errors. The expected \ref{eq:hybrid_td3_critic_loss} target can then be written as:
\begin{align}
\mathbb{E}[y]
&=\mathbb{E}[r]+\gamma\,\mathbb{E}\!\left[
\min\!\bigl(
Q^{\pi}(s',a')+\epsilon_{\phi_1},
Q^{\pi}(s',a')+\epsilon_{\phi_2}
\bigr)
\right] \nonumber \\
&=\mathbb{E}[r]+\gamma\Bigl(\mathbb{E}[Q^{\pi}(s',a')]+\mathbb{E}[\min(\epsilon_{\phi_1},\epsilon_{\phi_2})]\Bigr),
\end{align}
where the second equality follows from linearity of expectation. Defining the target bias as: 
$Bias
= \mathbb{E}[y] - Q^{\pi}_{\text{true}} = \mathbb{E}[\min(\epsilon_{\phi_1},\epsilon_{\phi_2})] = \mu-\tfrac{\sigma}{\sqrt{\pi}}$.

\noindent
\textit{(a) Hybrid TD3.}
$\delta Q_{\mathrm{Hybrid\_TD3}}=\mathbb{E}[\sum_{k=1}^{K} p_d(k) \min(\epsilon_{\phi_1},\epsilon_{\phi_2})] = \mathbb{E}[m_k]= \mu-\tfrac{\sigma}{\sqrt{\pi}}.$

\noindent
\textit{(b) HyDATD3.}
Let $\eta_j=\sum_k p_d(k)\min(\epsilon_{\phi_1}^{j,k},\epsilon_{\phi_2}^{j,k})$
with $\mathbb{E}[\eta_j]=-\tfrac{\sigma}{\sqrt{\pi}}$ and
$\mathrm{Var}(\eta_j)=\tau^2=\sigma^2(1-\tfrac{1}{\pi})\sum_k p_d(k)^2$.
Taking the max over two independent samples and applying Lemma~1: 
$\mathbb{E}[\max(S_1,S_2)]=Q^{\pi}(s',a')+\mathbb{E}[\max(\eta_1,\eta_2)]=Q^{\pi}(s',a')+\delta{Q_{HyDATD3}}$. The bias is:
$\delta Q_{\mathrm{HyDATD3}}= \mu -\tfrac{\sigma}{\sqrt{\pi}}+\tfrac{\tau}{\sqrt{\pi}}$.
Since $0<\tau<\sigma$, $\delta Q_{\mathrm{TD3\_Hybrid}}<\delta Q_{\mathrm{HyDATD3}}$.

\noindent
\textit{(c) HyDARC.}
$\delta Q_{\mathrm{HyDARC}}
=\lambda\mathbb{E}[\min(\eta_1,\eta_2)]+(1-\lambda)\mathbb{E}[\max(\eta_1,\eta_2)]
=\mu-\tfrac{\sigma}{\sqrt{\pi}}+\tfrac{\tau}{\sqrt\pi}(1-2\lambda)$.
For $\lambda<0.5$: $\delta Q_{\mathrm{TD3\_Hybrid}}<\delta Q_{\mathrm{HyDARC}}<\delta Q_{\mathrm{HyDATD3}}$.

\noindent
\textit{(d) HyTQC and (e) HyACC.}
Under Assumption~1, each atom $Q_{(i)}=Q^\pi+\mu+\sigma Z_{(i)}$ where $Z_{(i)}$ is the $i$-th order statistic of $NM$ i.i.d.\ standard normals.
Using the Blom approximation $\mathbb{E}[Z_{(i)}]\approx\Phi^{-1}(\tfrac{i-0.375}{NM+0.25})$:
\begin{equation}
\begin{aligned}
    &\delta Q_{\mathrm{HyTQC}}=\mu+\sigma\,\mathbb{E}\!\left[\tfrac{1}{kN}\textstyle\sum_{i=1}^{kN}Z_{(i)}\right], \\
&\delta Q_{\mathrm{HyACC}}=\mu+\sigma\,\mathbb{E}\!\left[\tfrac{1}{kN-\beta}\textstyle\sum_{i=1}^{kN-\beta}Z_{(i)}\right].
\end{aligned}
\end{equation}
Table~\ref{tab:quantile_bias} reports numerical values for $N=5$, $M=25$. Removing additional atoms ($\beta=2$) makes HyACC strictly more negative than HyTQC, so $\delta Q_{\mathrm{HyACC}}\leq\delta Q_{\mathrm{HyTQC}}$. Both are less negative than $\delta Q_{\mathrm{TD3\_Hybrid}}=\mu-\tfrac{\sigma}{\sqrt\pi}$ because the truncated mean of $Z_{(i)}$ over $kN<NM$ atoms exceeds $-1/\sqrt{\pi}$ (the full-sample limit). Combining all five inequalities yields Eq.~\eqref{eq:bias_order}.

\begin{table}[h]
\centering
\caption{Bias coefficients for HyTQC ($\beta{=}0$) and HyACC
($\beta{=}2$), $N{=}5$, $M{=}25$.}
\label{tab:quantile_bias}
% \begin{tabular}{@{}cccc@{}}
\begin{tabularx}{0.48\textwidth}{
    >{\centering\arraybackslash}X
    >{\centering\arraybackslash}m{0.11\textwidth}
    >{\centering\arraybackslash}m{0.13\textwidth}
    >{\centering\arraybackslash}m{0.14\textwidth}
}
\toprule
$k$ & $kN$ (HyTQC) & $\delta Q_{\mathrm{HyTQC}}$ & $\delta Q_{\mathrm{HyACC}}$ ($kN{-}2$)\\
\midrule
20 & 100 & $-0.3460$ & $-0.3702$ \\
21 & 105 & $-0.2865$ & $-0.3107$ \\
22 & 110 & $-0.2245$ & $-0.2496$ \\
23 & 115 & $-0.1591$ & $-0.1858$ \\
24 & 120 & $-0.0877$ & $-0.1173$ \\
\bottomrule
\end{tabularx}
\end{table}

% \section*{Acknowledgment}
% This work was supported by JST SPRING, Japan Grant Number JPMJSP2102.

\bibliographystyle{IEEEtran}
\bibliography{ref}  % .bib
\end{document}